\PassOptionsToPackage{preprint}{icml2027}
\documentclass{article}

\usepackage{microtype}
\usepackage{graphicx}
\usepackage{booktabs}
\usepackage{amsmath,amssymb,amsthm,mathtools}
\usepackage{algorithm}
\usepackage{algorithmic}
\usepackage{hyperref}
\usepackage{icml2027}
\hypersetup{pdftitle={Square-Root Error Amplification in Proper Autoregressive Learning}}
\ificmlshowauthors
  \hypersetup{pdfauthor={Tao Jiang, Haoyu Zhang, Yunhe LI, Shaowei Cai}}
\else
  \hypersetup{pdfauthor={Anonymous Authors}}
\fi

\icmltitlerunning{Square-Root Error Amplification in Proper Autoregressive Learning}

\newtheorem{theorem}{Theorem}
\newtheorem{lemma}{Lemma}[section]
\theoremstyle{definition}
\newtheorem{definition}{Definition}
\newcommand{\E}{\mathbb E}
\newcommand{\Prob}{\mathbb P}
\newcommand{\R}{\mathbb R}
\newcommand{\cD}{\mathcal D}
\newcommand{\cI}{\mathcal I}
\newcommand{\h}{\mathsf h}
\newcommand{\TV}{\mathrm{TV}}
\newcommand{\KL}{\mathrm{KL}}
\newcommand{\OPT}{\mathrm{OPT}}
\newcommand{\tr}{\operatorname{tr}}
\newcommand{\ind}{\mathbf 1}
\newcommand{\ip}[2]{\langle #1,#2\rangle}
\newcommand{\Bdot}{B_{\mathrm{dot}}}
\newcommand{\Capx}{C_{\mathrm{apx}}}
\newcommand{\excess}{\alpha}
\newcommand{\exfrac}{\widehat\alpha}
\allowdisplaybreaks[1]

\begin{document}
\twocolumn[
\icmltitle{Square-Root Error Amplification in Proper Autoregressive Learning}
\icmlsetsymbol{equal}{*}
\begin{icmlauthorlist}
\icmlauthor{Tao Jiang}{equal,iscas,ucas}
\icmlauthor{Haoyu Zhang}{equal,ucsd}
\icmlauthor{Yunhe LI}{equal,cityu}
\icmlauthor{Shaowei Cai}{iscas,ucas}
\end{icmlauthorlist}
{\ificmlshowauthors
  \begin{center}\vskip -0.03in
  \footnotesize
  \textsuperscript{1}Institute of Software, Chinese Academy of Sciences, Beijing, China\\
  \textsuperscript{2}School of Computer Science and Technology, University of Chinese Academy of Sciences, Beijing, China\\
  \textsuperscript{3}Department of Mathematics, University of California, San Diego, La Jolla, CA, USA\\
  \textsuperscript{4}Department of Computer Science, City University of Hong Kong, Hong Kong\\[3pt]
  Correspondence to: Shaowei Cai \textless{}caisw@ios.ac.cn\textgreater{}.\\
  Emails: \texttt{jiangt@ios.ac.cn}, \texttt{haz053@ucsd.edu}, \texttt{uuen.li@my.cityu.edu.hk}, \texttt{caisw@ios.ac.cn}
  \end{center}
  \vskip -0.15in
\fi}
\icmlaffiliation{iscas}{Institute of Software, Chinese Academy of Sciences, Beijing, China}
\icmlaffiliation{ucas}{School of Computer Science and Technology, University of Chinese Academy of Sciences, Beijing, China}
\icmlaffiliation{ucsd}{Department of Mathematics, University of California, San Diego, La Jolla, CA, USA}
\icmlaffiliation{cityu}{Department of Computer Science, City University of Hong Kong, Hong Kong}
\icmlcorrespondingauthor{Shaowei Cai}{caisw@ios.ac.cn}
\icmlkeywords{Autoregressive Learning, Agnostic Learning, Hellinger Distance, Computational Lower Bounds}
\vskip 0.3in
]
\ificmlshowauthors
  \makeatletter
  \global\icml@noticeprintedtrue
  {\let\thefootnote\relax\footnotetext{\hspace*{-\footnotesep}\textsuperscript{*}Equal contribution.}}
  \makeatother
\else
  \printAffiliationsAndNotice{}
\fi

\begin{abstract}
Autoregressive models are trained token by token but used to generate entire sequences, and under misspecification, maximum likelihood can amplify the best-in-class squared Hellinger error by a factor linear in the sequence length \(H\). We study how much of this amplification is computationally unavoidable when the learned model must lie in the model class. For binary autoregressive linear models with bounded parameters and features, we give a polynomial-time proper learner with approximation factor \(O(\min\{d,\sqrt H\})\), where \(d\) is the parameter dimension, and we prove a finer oracle inequality that adapts to the structure of the misspecification. Conversely, assuming \(\mathrm{NP}\not\subseteq\mathrm{BPP}\), no polynomial-time proper learner achieves a factor of order \(o(\sqrt H)\) at fixed accuracy and confidence uniformly over the dimension. Since a constant factor is achievable without computational constraints, these results identify square-root amplification as the sharp computational price of efficient proper autoregressive learning.
\end{abstract}

\section{Introduction}
\label{sec:intro}

Autoregressive models factorize a distribution over sequences into next-token conditionals. They are trained by maximizing the likelihood of each token given its observed prefix, which is known as teacher forcing, but at deployment they condition on their own samples, so early errors can shift the prefixes the model sees and compound along the sequence. This mismatch, known as exposure bias, motivated scheduled sampling and sequence-level training \citep{bengio15,ranzato16}; in imitation learning it appears as compounding error, whose cost can grow quadratically with the horizon \citep{ross11,rajaraman20}. We study a different source of error growth, which arises from misspecification even when no shift between training and generated prefixes is involved.

Under realizability, maximum likelihood learns the joint law in squared Hellinger distance at a rate governed by the complexity of the class \citep{zhang06}, and for policy classes this rate does not degrade with the horizon \citep{foster24}. Under misspecification, the population likelihood maximizer is the projection of the target onto the class in joint Kullback--Leibler divergence, which, unlike bounded statistical distances, can assign disproportionately large weight to trajectories to which the model assigns small probability. Following \citet{rohatgi25}, we instead compare the learned model with the best model in the class in squared Hellinger distance \(\h^2\) between joint trajectory laws, a bounded distance that controls total variation and underlies the horizon-free theory under realizability. Given independent trajectories from an arbitrary target law \(P\), the learner should output a law \(\widehat P\) with
\[
  \h^2(P,\widehat P)\le\varepsilon+\Capx\cdot\OPT,
\]
where \(\varepsilon>0\) is an accuracy parameter that more data can drive to zero, \(\OPT\) is the smallest squared Hellinger distance between \(P\) and a model in the class, and \(\Capx\ge1\) is the approximation factor. This is an agnostic guarantee in the sense of \citet{kearns94}.

\begin{figure}[t]
\centering
\includegraphics{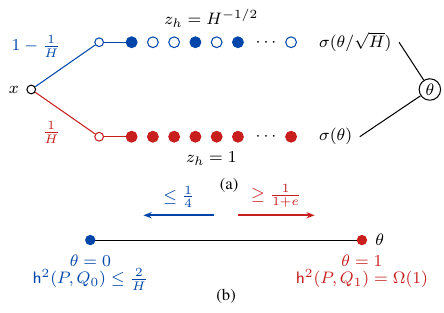}
\vskip -0.05in
\caption{The instance of Section~\ref{sec:likelihood-example}. (a) A context \(x\) is common (blue, probability \(1-1/H\)) or rare (red, probability \(1/H\)). Circles are tokens (filled for \(1\)); the blue row is one realization of fair bits. Both contexts share \(\theta\in[0,1]\). (b) On \([0,1]\), the common context's part of \(-\mathcal L'(\theta)\) pushes \(\theta\) toward \(0\) with magnitude at most \(1/4\), and the rare context's part pushes it toward \(1\) with magnitude at least \(1/(1+e)\), so the population likelihood is maximized at \(\theta=1\).}
\label{fig:likelihood}
\vskip -0.1in
\end{figure}

Under misspecification, maximum likelihood can amplify the approximation error even without exposure bias. When one parameter is shared across positions and contexts, a context of probability \(\varrho\) contributes at most \(2\varrho\) to the joint squared Hellinger distance but can contribute order \(\varrho H\) to the log loss, so likelihood fitting may sacrifice common contexts to fit rare ones. Figure~\ref{fig:likelihood} illustrates a scalar model with conditionally independent tokens and a rare context of probability \(1/H\) (Section~\ref{sec:likelihood-example}). Although a feasible parameter has squared Hellinger error \(O(1/H)\), the cumulative log loss of the rare context drives the population likelihood maximizer to a parameter with constant error. Likelihood fitting thus amplifies the best-in-class error by a factor \(\Theta(H)\), with infinite data and exact convex optimization.

Linear amplification is not information-theoretically necessary. A \(\rho\)-estimator \citep{baraud18}, which selects a member of the class, attains \(\Capx=O(1)\) but requires min-max optimization over the class \citep{rohatgi25}. Known polynomial-time methods, in contrast, achieve factors of order \(H\). \citet{rohatgi25} show that every learner that builds the model one position at a time, next-token prediction included, can incur \(\Capx=\Omega(H)\), and they analyze a robustified log loss that attains \(O(H)\) in polynomial time; their computational lower bound excludes only factors smaller than every fixed power of \(\max\{d,H\}\). The computational limits of learning within the class thus remained open.

We study this question for proper learners, which return a model from the class: here, a single feasible parameter shared by all positions and contexts, deployed with the model's own sampling procedure. Properness is a classical constraint in computational learning theory, where it can separate tractable from intractable problems \citep{pitt88}, and improper outputs can attain strictly better approximation factors in density estimation \citep{bousquet19}. Our central question is the following.
\begin{center}
\emph{What is the smallest worst-case approximation factor achievable by a polynomial-time proper learner?}
\end{center}

\subsection{Our contributions}
\label{sec:results}

\textbf{Efficient proper learning.} We give a polynomial-time algorithm that, from a number of trajectories linear in \(1/\varepsilon\), returns a single parameter of the class with approximation factor \(O(\min\{d,\sqrt H\})\) (Theorem~\ref{thm:proper}). This improves the linear horizon dependence of likelihood fitting to a square root, and with \(c=1/2\) it answers the question of \citet[Section~5.2]{rohatgi25} whether efficient learners can achieve \(\Capx=H^c\) for some \(c<1\).

\textbf{Adaptation to misspecification.} The guarantee is an oracle inequality against every comparator in the class, in which the factor \(\min\{d,\sqrt H\}\) multiplies only the \emph{excess mass}, the amount by which the target density exceeds twice the comparator density. Consequently, if the target density is at most twice the density of some comparator, the bound is a constant multiple of that comparator's error. On the instance of Figure~\ref{fig:likelihood} the bound is \(\varepsilon+O(1/H)\) for a different reason: there \(d=1\), and the comparator \(\theta=0\) has error and excess mass both of order \(1/H\).

\textbf{Matching computational lower bound.} Assuming \(\mathrm{NP}\not\subseteq\mathrm{BPP}\), no polynomial-time proper learner achieves the factor \(c\sqrt H\) at fixed accuracy and confidence uniformly over the dimension, and the same holds for \emph{context-independent mixtures}, which draw a parameter independently of the context and use it for the whole sequence (Theorem~\ref{thm:lower}). Together with Theorem~\ref{thm:proper}, this shows that the optimal dimension-uniform worst-case factor of efficient proper learning is of order \(\sqrt H\).

\subsection{Technical overview}
\label{sec:overview}

The upper bound must control trajectories that the comparator fits poorly while keeping the likelihood curvature that drives optimization. Along a difference of feasible parameters, the gradient of the negative log likelihood of a trajectory, which we call its score, can have projection of order \(H\), so poorly fitted trajectories carrying excess mass \(\excess\) can turn the empirical gradient away from the comparator. Our learner reweights trajectories under a second-moment constraint on the weighted scores, with a logarithmic objective that makes removing a few exceptional trajectories cheap and removing many typical ones costly. The reweighted gradient still leaves an error of order \(\excess/\tau\) on the exceptional trajectories, which a multiple \(\tau\) of the unweighted gradient absorbs by restoring curvature, at the price \(\tau H\excess\). Balancing \(\excess/\tau+\tau H\excess\) at \(\tau=H^{-1/2}\) produces the square root. When \(d\le\sqrt H\), likelihood fits on unions of data blocks take the place of reweighting.

The lower bound turns the consistency of a shared parameter into a Label Cover constraint. A feasible parameter assigns soft labels to the vertices of a graph, and matching an efficiently samplable target forces the soft labels at the endpoints of most edges to agree. On graphs of low value this fails, and every parameter has a logit error of order \(1/m\) on a constant fraction of contexts, while \(\OPT=O(1/m)\) on satisfiable graphs. With \(H\asymp m^2\) tokens, such an error shifts the expected number of ones by order \(\sqrt H\), comparable to its standard deviation, so a single trajectory detects it with constant statistical power, and absorbing the error requires a factor of order \(m\asymp\sqrt H\).

\subsection{Related work}

Whether exposure bias matters in practice, and whether modified training objectives address it, has been debated \citep{huszar15,he21,arora22}. Beyond the results of \citet{rohatgi25} recalled above, they give a kernel relaxation for binary autoregressive linear models that attains the factor \(H/K\) in time growing with \(K\), and a noisy-parity bound that excludes, even for improper learners, factors of order \(\exp(\log^{1-c}\max\{d,H\})\). Our lower bound instead concerns proper learners and context-independent mixtures and excludes \(c\sqrt H\). The horizon-free guarantees of \citet{foster24} concern log-loss behavior cloning under realizability, where \(\OPT=0\). \citet{li24} study agnostic interactive imitation with expert queries, \citet{xu26} prove oracle inequalities against a joint KL comparator, which can exceed the Hellinger benchmark by a factor growing with \(H\), and \citet{blondel26} study sequence-level objectives evaluated on model-decoded sequences.

Our validation step uses the pairwise comparison of \(\rho\)-estimation \citep{baraud18}. Robust estimators for logistic regression and generalized linear models tolerate adversarial or oblivious corruption of a fraction of samples \citep{feng14,mukhoty23,diakonikolas23glm}, and \citet{mourtada22} study excess log loss for misspecified logistic regression. Filtering methods remove samples along directions of excessive empirical second moment \citep{diakonikolas19robust}, also when applied to gradients \citep{prasad20,diakonikolas19sever}. In our setting the exceptional trajectories arise from misspecification rather than corruption, and the weights must also preserve the likelihood curvature of the typical trajectories. On the computational side, agnostic learning can be harder than realizable learning \citep{kearns94}, and our reduction starts from the Label Cover hardness of \citet{bhangale26}.

\section{Problem formulation and main results}
\label{sec:model}

\subsection{Model, data and error}

Let \(H,d\ge1\). A trajectory \(o=(x,y_{1:H})\) consists of a context \(x\) and a token sequence \(y_{1:H}\in\{0,1\}^H\). For a known feature map \(\phi\), the contrast at the prefix \((x,y_{<h})\) is
\[
 z_h=\phi(x,y_{<h}1)-\phi(x,y_{<h}0).
\]
A parameter \(\theta\) in a nonempty closed convex set \(\Theta\subseteq\R^d\) predicts the next token with probability \(p_{\theta,h}=\sigma(\ip{\theta}{z_h})\), where \(\sigma(t)=(1+e^{-t})^{-1}\), so that its conditional sequence law is
\[
 q_\theta(y_{1:H}\mid x)=
      \prod_{h=1}^H p_{\theta,h}^{y_h}(1-p_{\theta,h})^{1-y_h}.
\]
The same parameter is used at every context and position. We assume that, for all feasible parameters and all action features,
\[
 \|\theta\|_2,\ \|\phi\|_2\le B=\sqrt d,\qquad
                  |\ip{\theta}{\phi}|\le \Bdot=1 .
\]
Feasible logits therefore lie in \([-2,2]\), where \(\sigma'\) is at least \(\kappa=\min_{|t|\le2}\sigma'(t)>0\). These are the norm bounds of \citet[Theorem~5.3]{rohatgi25}. The learner accesses \(\phi\) and the Euclidean projection onto \(\Theta\) through oracles, and our algorithms make polynomially many oracle calls at inverse-polynomial precision.

The target \(P\) is an arbitrary law on trajectories with an unknown context marginal \(\cD\). The learner observes independent trajectories from \(P\) and outputs a policy with efficient conditional generation. Each output induces a joint law with marginal \(\cD\); the \emph{model law} \(Q_\theta\) has conditional law \(q_\theta\). With densities taken with respect to \(\cD\) times counting measure, \(Q_\theta\) has density \(q_\theta(y_{1:H}\mid x)\) at \(o\), and \(P\) has density \(p\). We measure error by the unhalved squared Hellinger distance \(\h^2(Q,Q')=\int(\sqrt q-\sqrt{q'})^2\in[0,2]\) between laws \(Q,Q'\) with densities \(q,q'\), and write \(\h\) for the Hellinger distance. For a comparator \(v\in\Theta\), let
\[
 \Delta_v=\h^2(P,Q_v),\ \ \OPT=\inf_{v}\Delta_v,\ \ \excess_v=\textstyle\int(p-2q_v)_+ .
\]
The excess mass \(\excess_v\) is the total amount by which \(p\) exceeds \(2q_v\); it is at most, and in general smaller than, \(P(p>2q_v)\).

\begin{definition}[Output classes]\label{def:outputs}
A learner is \emph{proper} if it outputs \(Q_{\widehat\theta}\) for some \(\widehat\theta\in\Theta\). A \emph{context-independent mixture} draws \(v\in\Theta\) independently of a new context \(x\) and then generates the entire sequence from \(Q_v(\cdot\mid x)\). An \emph{unrestricted} learner may output any policy with efficient conditional generation.
\end{definition}

\subsection{Main results}

\begin{theorem}[An excess-mass oracle inequality for efficient proper learning]\label{thm:proper}
Let \(0<\varepsilon,\delta<1/2\). Algorithm~\ref{alg:proper} uses
\[
 n=O\left(\frac{d\log(2+H)+\min\{d,\sqrt H\}\log(2/\delta)}
                  {\varepsilon}\right)
\]
independent trajectories from \(P\), runs in time polynomial in \(d,H,1/\varepsilon\) and \(\log(2/\delta)\) with the feature and projection oracles, and returns \(\widehat\theta\in\Theta\) such that, with probability at least \(1-\delta\),
\[
  \h^2(P,Q_{\widehat\theta})\le\varepsilon+
       C\inf_{v\in\Theta}\bigl\{\Delta_v+\min\{d,\sqrt H\}\,\excess_v\bigr\},
\]
where \(C\) is a universal constant. Drawing a token from \(Q_{\widehat\theta}\) costs two feature evaluations, one inner product and one Bernoulli draw.
\end{theorem}

Since \((p-2q_v)_+\le2(\sqrt p-\sqrt{q_v})^2\) pointwise, \(\excess_v\le2\Delta_v\), and Theorem~\ref{thm:proper} gives
\[
 \h^2(P,Q_{\widehat\theta})\le\varepsilon+C'\min\{d,\sqrt H\}\,\OPT
\]
for a universal \(C'\). The dependence on the horizon enters only through the coefficient of the excess mass. For instance, if \(P=(1-\varrho)P_0+\varrho P_1\) with \(P_0\le2Q_v\), then \(\excess_v\le\varrho\), so the factor is charged only to the component that the comparator cannot explain.

\begin{theorem}[Computational lower bound for proper learning]\label{thm:lower}
Assume \(\mathrm{NP}\not\subseteq\mathrm{BPP}\). There are universal constants \(c>0\), \(\varepsilon_0\in(0,1/2)\) and \(C_0\in\mathbb N\) such that, for every integer \(m\ge2\) and \(H=C_0m^2\), the following holds. No randomized algorithm that runs in time polynomial in \(d\) and outputs either a parameter \(\widehat\theta\in\Theta\) or an efficiently samplable context-independent mixture with law \(\widehat R\) guarantees, with probability at least \(9/10\) on every instance with horizon \(H\),
\[
 \h^2(P,\widehat R)\le \varepsilon_0+c\sqrt H\,\OPT ,
\]
where \(\widehat R=Q_{\widehat\theta}\) for a proper output.
\end{theorem}

Both theorems concern the same model, access, error measure and comparator. At accuracy \(\varepsilon_0\) and confidence \(9/10\), Theorem~\ref{thm:proper} attains \(C'\sqrt H\) in every dimension, while at each \(H=C_0m^2\) no efficient proper learner or context-independent mixture attains \(c\sqrt H\) in every dimension. The optimal dimension-uniform factor of efficient proper learning is therefore \(\Theta(\sqrt H)\); the regime \(d\ll\sqrt H\) and unrestricted outputs remain open (Section~\ref{sec:discussion}).

\subsection{Linear amplification of likelihood fitting}
\label{sec:likelihood-example}

Let \(H\ge2\), \(d=1\), \(\Theta=[0,1]\), and let the action features be \(\pm z_h/2\). A rare context of probability \(1/H\) has all contrasts equal to \(1\) and the all-ones target sequence. The common context has contrasts \(1/\sqrt H\) and fair target bits. Then \(\OPT\le\Delta_0\le2/H\), where \(\Delta_0\) is the error of \(\theta=0\). Since every model law gives the rare sequence probability at most \(\sigma(1)^H\), also \(\OPT\ge2(1-\sigma(1)^{H/2})/H\), so \(\OPT\) is of order \(1/H\). The population log loss \(\mathcal L(\theta)=\E_P[-\log q_\theta]\) has derivative
\[
 \mathcal L'(\theta)
  = (1-1/H)\sqrt H
                 \bigl(\sigma(\theta/\sqrt H)-1/2\bigr)
    +\sigma(\theta)-1 .
\]
On \([0,1]\), the first term is at most \(1/4\) by smoothness of \(\sigma\) and the second is at most \(-1/(1+e)<-1/4\), so the likelihood is maximized at \(\theta=1\). At \(\theta=1\), the mean of each common token exceeds \(1/2\) by at least \(\kappa/\sqrt H\), so each one-token squared Hellinger distance is at least \(\kappa^2/H\). Multiplying the \(H\) one-token affinities gives conditional squared Hellinger distance at least \(2(1-e^{-\kappa^2/2})\) on a context of probability at least \(1/2\). The likelihood fit thus pays a factor \(\Theta(H)\). By the same bound \(\sigma(1)^H\), every parameter has joint KL divergence at least \(\log(1/\sigma(1))\) from \(P\), so the best-in-class joint KL divergence is a constant while \(\OPT=\Theta(1/H)\). In contrast, \(\excess_0\le1/H\) because \(Q_0\) agrees with \(P\) on the common context, and Theorem~\ref{thm:proper} with \(d=1\) gives squared Hellinger error \(\varepsilon+O(1/H)\).

\section{Efficient proper autoregressive learning}
\label{sec:proper}

\begin{algorithm}[t]
\caption{Localized proper learning}
\label{alg:proper}
\begin{algorithmic}[1]
\REQUIRE accuracy \(\varepsilon\), confidence \(\delta\); samples from \(P\); feature and projection oracles
\STATE \(L_H\leftarrow\log(2+H)\), \(\gamma_H\leftarrow\min\{d,\sqrt H\}\), \(J\leftarrow\lceil C_JL_H\rceil\), \(r_j\leftarrow2^{2-j}\), \(J_\varepsilon\leftarrow\min\{j:r_j\le c_\varepsilon\sqrt\varepsilon\}\), \(\iota_j\leftarrow\log(C/\delta)+C(J_\varepsilon-j+1)\); choose any \(a_1\in\Theta\)
\FOR{\(j=1\) {\bfseries to} \(J_\varepsilon-1\)}
  \STATE draw fresh pilot, fitting and validation samples of size \(O((dL_H+\gamma_H\iota_j)/r_j^2)\) each
  \STATE \(\widehat M\leftarrow\) pilot average of the design matrices \(A\) of Section~\ref{sec:filter}; form \(K_{r_j}\) and \(M_0\) by \eqref{eq:algorithm-pilot}
  \IF{\(d>\sqrt H\)}
    \STATE run \(T=\mathrm{poly}(d,H,1/\varepsilon)\) projected steps \(w\leftarrow\Pi_{K_{r_j}}(w-\eta\, g(w))\), \(\eta=1/(3H\sqrt T)\), with the filtered gradient \(g\) of \eqref{eq:program}; let \(\mathcal C\) be \(C\iota_j\) uniformly drawn iterates
  \ELSE
    \STATE split the fitting sample into \(C(1+\lceil\iota_j/J\rceil)\) pools of \(4J\) blocks; let \(\mathcal C\) be the likelihood fits over \(K_{r_j}\) on all unions of \(J\) blocks of a pool
  \ENDIF
  \STATE anchor \(a_{j+1}\leftarrow\textsc{Select}(\mathcal C)\) on the validation sample, uniform weights
\ENDFOR
\STATE \textbf{return} \(\widehat\theta\leftarrow\textsc{Select}(a_1,\dots,a_{J_\varepsilon})\) on \(C\log(2/\delta)/\varepsilon\) fresh trajectories, weights \(\pi_j\propto2^{j-J_\varepsilon}\)
\end{algorithmic}
\end{algorithm}

Algorithm~\ref{alg:proper} proceeds through a sequence of localization stages with radii \(r_j=2^{2-j}\). Stage \(j\) builds an ellipsoid \(K_{r_j}\) around the current \emph{anchor}, the parameter returned by the previous stage, from an independent pilot sample. It then produces a short list of candidate parameters inside the ellipsoid from a fitting sample, and the candidate selected by pairwise likelihood-ratio comparisons on a validation sample becomes the next anchor. Throughout, \(L_H=\log(2+H)\) and \(\gamma_H=\min\{d,\sqrt H\}\). For candidate parameters \(\theta_1,\dots,\theta_K\), weights \(\pi_k>0\) summing to one and \(n_{\rm val}\) validation trajectories, \(\textsc{Select}\) returns the \(\theta_k\) that minimizes \(t_k+\max_l(\overline T_{kl}-t_l)\), where \(t_k=C\log(2/\pi_k)/n_{\rm val}\) and \(\overline T_{kl}\) is the validation average of
\[
 T_{kl}(o)=\frac{\sqrt{q_{\theta_l}(o)}-\sqrt{q_{\theta_k}(o)}}
                    {\sqrt{q_{\theta_l}(o)}+\sqrt{q_{\theta_k}(o)}} .
\]
The statistic is bounded, and the unknown context density cancels from it.

The analysis fixes a comparator \(v\in\Theta\) and follows the anchors. A stage of radius \(r\) must return an anchor whose model law is within Hellinger distance \(r/2\) of \(Q_v\) whenever
\begin{equation}
      \h(Q_a,Q_v)\le r,\qquad
      \Delta_v+\gamma_H\excess_v\le c_*r^2 ,
 \label{eq:main-stage}
\end{equation}
that is, when the anchor \(a\) is close and the comparator's error is small at scale \(r\). When \(d>\sqrt H\), this reduces to the uniform lower bound \eqref{eq:direction-main} on the progress \(\ip{w-v}{g(w)}\) of the filtered gradient over the ellipsoid. Sections~\ref{sec:score-likelihood}--\ref{sec:inputs} establish this bound, and Appendix~\ref{app:proper} contains the complete proof.

\subsection{Likelihood evidence from moderate-density trajectories}
\label{sec:score-likelihood}

Write \(h_w=\h(Q_v,Q_w)\) for the Hellinger distance between model laws and \(\cI=\{p\le4q_v\}\) for the set of trajectories on which the target density exceeds the comparator density by at most a factor of four. Since \(p-2q_v>p/2\) off \(\cI\), the complement has probability \(P(\cI^c)\le2\excess_v\). On \(\cI\) we use the capped log-likelihood ratio
\[
          Y_w=\ind_\cI\min\{12,\log(q_v/q_w)\},
\]
which satisfies, at each fixed candidate \(w\),
\begin{equation}
               \E_P e^{-Y_w/4}\le e^{-h_w^2/8+C\Delta_v}.
 \label{eq:likelihood-moment}
\end{equation}
To see this, put \(s=q_w/q_v\) and \(T=\max\{s^{1/4},e^{-3}\}-1\), so that \(e^{-Y_w/4}=1+\ind_\cI T\). The arithmetic--geometric mean inequality gives \(T\le(\sqrt s-1)/2\), and also \(|T|\le|\sqrt s-1|\), so \(\E_{Q_v}T\le-h_w^2/4\) and \(\E_{Q_v}T^2\le h_w^2\). On \(\cI\) we have \(|p-q_v|\le3\sqrt{q_v}\,|\sqrt p-\sqrt{q_v}|\), and changing measure from \(Q_v\) to \(P\) there costs at most \(3\sqrt{\Delta_v}\,h_w\) by Cauchy--Schwarz. Restricting to \(\cI\) costs at most \(Q_v(\cI^c)\le\Delta_v\), because \(T\ge-1\) and \(p>4q_v\) off \(\cI\). Young's inequality then gives \eqref{eq:likelihood-moment}. The calculation holds for any logistic parameter \(w\), feasible or not, which allows averaging over small perturbations in Section~\ref{sec:inputs}.

By Jensen's inequality, \eqref{eq:likelihood-moment} yields \(\E_PY_w\ge h_w^2/2-O(\Delta_v)\), so the capped ratio detects Hellinger separation up to the comparator's error. The event \(\cI\) enters only the analysis.

\subsection{A second-moment-constrained gradient}
\label{sec:filter}

For a trajectory \(o\), the score and the design matrix are
\[
 G_w(o)=\sum_h z_h(\sigma(\ip{w}{z_h})-y_h),\qquad
 A(o)=\sum_h z_hz_h^\top ,
\]
so \(G_w=-\nabla_w\log q_w(o)\) is the gradient of the negative log likelihood, and \(u^\top A(o)u\), the \emph{energy} of a direction \(u\), is the sum of the squared logit changes along \(o\). For \(u=w-v\) and \(V=u^\top A(o)u\), the curvature of the logistic loss on feasible logits gives
\[
 \ip{u}{G_w(o)}\ge\log(q_v(o)/q_w(o))+\kappa V/2 .
\]
A few trajectories with large scores can keep the average score from pointing away from \(v\), even when \(w\) is far from \(v\). To control them, the pilot sample gives the average \(\widehat M\) of its design matrices, and at radius \(r\) and anchor \(a\) the learner uses the ellipsoid and the ridge
\begin{equation}
 \begin{aligned}
 K_r&=\{w\in\Theta:\|w-a\|_{\widehat M}\le C_K\sqrt H\,r\},\\
 M_0&=\frac{\lambda_*}{1+H}(\widehat M+\lambda I_d),
      \qquad \lambda=\frac{r^2}{16(1+B)^2},
 \end{aligned}
 \label{eq:algorithm-pilot}
\end{equation}
where \(\|u\|_M=\sqrt{u^\top Mu}\), \(C_K\) is a large constant and \(\lambda_*\) a small one. With \(n\) fitting trajectories, empirical average \(P_n\), average design matrix \(\widehat A=P_nA\) and \(\tau=H^{-1/2}\), the learner computes at a candidate \(w\in K_r\) an approximate solution of
\begin{equation}
 \begin{aligned}
 &\text{maximize} && P_n\log\frac{\omega_i+\tau}{1+\tau}\\
 &\text{subject to} &&
 P_n\omega_i^2 G_wG_w^\top\preceq C(\widehat A+M_0),\\
 &&& 0\le \omega_i\le1 ,
 \end{aligned}
 \label{eq:program}
\end{equation}
and moves along the \emph{filtered gradient} \(g(w)=P_n\omega_iG_w+2\tau P_nG_w\), the weighted score average plus \(2\tau\) times the unweighted one. The constraint bounds the uncentered second-moment matrix of the weighted scores, not a covariance, and \eqref{eq:program} is a convex program that top-eigenvector subgradients solve to the required precision (Appendix~\ref{app:filtered}).

\subsection{Descent from the constrained program}
\label{sec:retention}

Fix a stage with radius \(r\), fitting size \(n\) and confidence parameter \(\iota\ge1\), and a candidate \(w\in K_r\). Write \(u=w-v\), \(V_i=u^\top A_iu\), \(X=P_nV_i\), and \(\exfrac=P_n\ind_{\cI^c}\) for the empirical excess fraction. On \(K_r-K_r\) the ridge satisfies \(u^\top M_0u\le\zeta r^2\) for a small fixed \(\zeta>0\). We need a lower bound on \(\ip{u}{g(w)}\) in terms of \(h_w^2\). By the curvature inequality of Section~\ref{sec:filter}, each trajectory satisfies \(\ip{u}{G_w(o_i)}\ge\ell_i+(\kappa/2)V_i\), where \(\ell_i=\log(q_v/q_w)(o_i)\). Two difficulties remain: the optimal weights might discard the trajectories that carry the likelihood evidence, and on trajectories outside \(\cI\) the log ratio \(\ell_i\) is not controlled. The argument uses two statistical inputs, both proved in Appendix~\ref{app:proper}. The first, a witness (W), is a set \(\cI_0\subseteq\cI\) of fitting trajectories, which omits at most \(C(d+\iota)\) trajectories of high leverage relative to the pilot design matrices, together with weights \(0<y_i\le1\) on \(\cI_0\) satisfying
\[
 \begin{aligned}
 P_n\ind_{\cI_0}y_i^2G_vG_v^\top
   &\preceq C(\widehat A+M_0),\\
 P_n\ind_{\cI_0}(-\log y_i)
   &\le C_{\lambda_*}\{\Delta_v+(dL_H+\iota)/n\}.
 \end{aligned}
\]
The second, a local likelihood bound (L), is the uniform version \eqref{eq:local-likelihood} of the test of Section~\ref{sec:score-likelihood}. Simultaneously for all \(w\in K_r\),
\[
  P_nY_w\ge c_1 h_w^2-C\Delta_v-\zeta r^2-C_\zeta(dL_H+\iota)/n .
\]

The witness controls the comparator scores \(G_v\), whereas the program constrains the scores \(G_w\) at the candidate, so we first transport it to \(w\). Since \(\sigma\) is \(1/4\)-Lipschitz, Cauchy--Schwarz gives, for every direction \(e\),
\[
   \bigl\langle e,G_w-G_v\bigr\rangle^2
                      \le (V_i/16)\,e^\top A_i e ,
\]
so the scores move little except on trajectories with large energy \(V_i\). Shrinking the weights of these trajectories compensates for the move. With \(a_0=\kappa/24\), consider the weights \(y_i/(1+a_0V_i)\) on \(\cI_0\) and zero elsewhere. Writing \(G_w=G_v+(G_w-G_v)\) and using \(V_i/(1+a_0V_i)^2\le1/(4a_0)\), these weights satisfy the constraint of \eqref{eq:program} at \(w\) for a large enough absolute multiplier.

Optimality of the weights \(\omega_i\) can now be compared with this feasible point. Write \(s_\tau(t)=(t+\tau)/(1+\tau)\) and \(L_\tau=\log(1+1/\tau)\). The feasible point has log value at least \(\log y_i-\log(1+a_0V_i)\) on \(\cI_0\) and exactly \(-L_\tau\) elsewhere, so optimality and \(\log t\le t-1\) give
\[
 \begin{aligned}
 &P_n\ind_\cI[\omega_i(1+a_0V_i)-1]\\
 &\quad\ge-\tau a_0X-C L_\tau \exfrac-C\zeta r^2\\
 &\qquad\quad-C_\zeta\{\Delta_v+(d+\iota)L_H/n\}.
 \end{aligned}
\]
The first term comes from \(\omega_i=(1+\tau)s_\tau(\omega_i)-\tau\), and the others come from the omitted trajectories, each costing at most \(L_\tau\), from the witness cost and from the optimization tolerance. In words, the optimal weights cannot be much smaller than one on \(\cI\), once high-energy trajectories are discounted. On \(\cI\), since \(\omega_i\le1\), we have \(\omega_i\ell_i\ge\min(12,\ell_i)-12(1-\omega_i)\), and adding the weighted curvature \((\kappa/2)\,\omega_iV_i=12a_0\,\omega_iV_i\) gives
\[
 \omega_i\bigl(\ell_i+\tfrac{\kappa}{2}V_i\bigr)\ge Y_w(o_i)+12\bigl[\omega_i(1+a_0V_i)-1\bigr].
\]
Averaging over \(\cI\) and applying the previous display, the weighted gradient retains the capped evidence \(P_nY_w\) up to the stage error, at the price of the term \(-12\tau a_0X=-(\kappa/2)\tau X\).

Two costs arise off \(\cI\), and they determine the choice of \(\tau\). There the log ratio is uncontrolled, but the constraint and Cauchy--Schwarz bound the weighted gradient in direction \(u\) by \(C\sqrt{\exfrac(X+\zeta r^2)}\). This square root can be absorbed only by a curvature term proportional to \(X\), which the unweighted gradient supplies on all trajectories. Since \(V_i\le16H\) and \(|\ip{u}{G_w}|\le4H\) on every trajectory,
\[
    \langle u,P_nG_w\rangle\ge P_nY_w+(\kappa/2)X-CH\exfrac .
\]
Adding \(2\tau\) times the unweighted gradient therefore contributes \(\kappa\tau X\), which covers the term \(-(\kappa/2)\tau X\) above and leaves a positive multiple of \(\tau X\). Young's inequality then gives
\[
 C\sqrt{\exfrac(X+\zeta r^2)}\le\tfrac{\kappa}{4}\tau X+C'\exfrac/\tau+\zeta r^2 .
\]
The first cost, \(\exfrac/\tau\), comes from absorbing the uncontrolled weighted gradient; the second, \(\tau H\exfrac\), is what the unweighted gradient pays off \(\cI\), where its projection can be as large as \(4H\). These costs balance at \(\tau=H^{-1/2}\), where each equals \(\sqrt H\,\exfrac\). Finally, a Bernoulli count gives \(\exfrac\le C(\excess_v+\iota/n)\) with high probability. Together with (W), (L) and \(L_\tau,L_H=O(\sqrt H)\), this yields, simultaneously for all \(w\in K_r\),
\begin{equation}
 \begin{aligned}
 \ip{w-v}{g(w)}&\ge c_2\,h_w^2\\
 &\quad-C(\Delta_v+\sqrt H\,\excess_v+\zeta r^2)\\
 &\quad-C_\zeta(dL_H+\sqrt H\,\iota)/n.
 \end{aligned}
 \label{eq:direction-main}
\end{equation}
Because the witness pays only a logarithmic price for the exceptional trajectories, the dimension term carries \(\log(2+H)\) rather than \(\sqrt H\), and the factor \(\sqrt H\) appears only in the excess and confidence terms.

\subsection{Constructing the witness and localizing}
\label{sec:inputs}

The witness (W) must show that the comparator scores satisfy the second-moment constraint after a reweighting of small logarithmic cost. A matrix inequality holds when it holds in every direction, so the proof starts from a single direction \(t\). Under \(Q_v\), the projected score \(\ip{t}{G_v}\) is a martingale along the sequence whose increments lie in ranges of length \(|\ip{t}{z_h}|\), also when the contrasts depend on earlier tokens. Conditional Hoeffding bounds therefore control the exponential moment of \(\log(1+\ip{t}{G_v}^2/2)-t^\top At/8\): a trajectory can have a large projected score only if it has a comparably large energy in direction \(t\). The change of measure of Section~\ref{sec:score-likelihood} transfers this bound to \(P\) on \(\cI\).

A union bound over directions would cost a factor depending on the precision. Instead, the proof integrates the directional tests against a Gaussian prior on \(t\) defined by the ridge and applies the entropy variational inequality, which gives a bound for every Gaussian posterior at the price of its relative entropy to the prior. A posterior with covariance \(2\Lambda\) plus a baseline makes \(\ip{t}{G_v}^2\) comparable to \(G_v^\top\Lambda G_v\) on each trajectory, which turns the directional tests into control of \(P_n\ind_{\cI_0}\log(1+G_v^\top\Lambda G_v)\). The relative entropy is governed by a log-determinant of the fitting design normalized by the pilot ridge, and this is where leverage matters: a fitting trajectory whose design matrix is large relative to the pilot would inflate the determinant. A trace identity bounds the leverage of fresh trajectories without any eigenvalue condition on the population design (Lemma~\ref{lem:leverage}), so at most \(C(d+\iota)\) trajectories are discarded and the entropy cost is \(O(dL_H+\iota)\). The resulting bound controls \(P_n\ind_{\cI_0}\log(1+G_v^\top\Lambda G_v)\) by \(C\tr(\Lambda(\widehat A+M_0))\) plus the stage error, for every positive semidefinite \(\Lambda\).

This family of inequalities is the dual form of (W). Minimizing the logarithmic cost of the weights subject to the second-moment constraint is a convex program, and its Lagrange dual with a matrix multiplier \(\Lambda\) charges each trajectory
\[
 \inf_{0<y\le1}\{-\log y+(G_v^\top\Lambda G_v)\,y^2\}\le2\log(1+G_v^\top\Lambda G_v).
\]
The dual cost of a trajectory is thus at most logarithmic in its score, the bound above pays for it, and strong duality yields weights satisfying (W) (Appendix~\ref{app:witness}). The same calculation explains the logarithmic objective of \eqref{eq:program}: a trajectory with a large score can be down-weighted at a cost that is only logarithmic in its score.

Localization uses the fact that Hellinger separation controls energy. Let \(L_h=q_{w,\le h}/q_{v,\le h}\) be the likelihood ratio of prefixes and \(h_h^2\) the conditional squared Hellinger distance of the next-token laws. Telescoping the joint affinity gives
\[
            h_w^2=\textstyle\sum_h \E_{Q_v}\sqrt{L_{h-1}}\,h_h^2 .
\]
Before the first prefix with \(L_h<1/4\), each weight is at least \(1/2\), and \(h_h^2\ge\kappa^2\ip{w-v}{z_h}^2\). By data processing applied at that stopping time, the probability of reaching it is at most \(4h_w^2\). Hence
\begin{equation}
          \E_{Q_v}\min\{1,(w-v)^\top A(w-v)\}\le C h_w^2 .
 \label{eq:energy-capped}
\end{equation}
Since the energy of feasible differences never exceeds \(16H\), \eqref{eq:energy-capped} places \(v\) in \(K_r\) whenever \eqref{eq:main-stage} holds, even when the design adapts to earlier tokens. Localization also keeps the uniform test cheap: in the norm \(\|u\|_{M_P}+r\|u\|_*\), with \(M_P=\E_PA\) and \(\|u\|_*=\sup(u^\top Au)^{1/2}\) over trajectories, \(K_r\) has radius \(O(\sqrt H\,r)\), and to make the test of Section~\ref{sec:score-likelihood} uniform over \(K_r\), the proof averages the capped log ratio over a symmetric ball of radius \(cr/\sqrt H\) around each candidate. Symmetry cancels the linear term, and since the two radii differ by a factor \(O(H)\), the entropy cost is \(O(d\log(2+H))\), with no covering at precision \(\varepsilon\) (Appendix~\ref{app:local}).

\subsection{Small dimension, validation and the proof of Theorem~\ref{thm:proper}}
\label{sec:schedule}

When \(d\le\sqrt H\), condition \eqref{eq:main-stage} has \(\gamma_H=d\), and the learner fits the likelihood on blocks of data instead of filtering. The normalization \(P_0\) of \(\min(p,2q_v)\) is a mixture component of \(P\) of mass \(1-\excess_v\), and \(\chi^2(P_0,Q_v)\le C\Delta_v\), so the high-probability guarantee of a likelihood fit over \(K_r\) on samples from \(Q_v\) transfers to samples from \(P_0\). In this latent decomposition, call a block of \(O(d/r^2)\) trajectories \emph{clean} if all of them come from \(P_0\); the learner never observes which blocks are clean. Each block is clean with probability at least \(3/4\) when \(d\excess_v\le c_*r^2\), so with probability \(1-e^{-\Omega(J)}\) at least \(J\) of \(4J\) blocks are clean, and the fit on their union is then close to \(Q_v\) with high probability. Since the learner fits all \(\binom{4J}{J}\le2^{4J}=\mathrm{poly}(H)\) unions, its list contains such a fit with high probability (Appendix~\ref{app:small}).

In both branches the candidate list is independent of the validation sample. Under \(Q_{\theta_k}\) the statistic \(T_{kl}\) has negative mean of order \(\h^2(Q_{\theta_k},Q_{\theta_l})\), and this persists under \(P\) when \(Q_{\theta_k}\) is close to \(P\), so \(\textsc{Select}\) returns \(\theta_{\rm out}\) with \(\h^2(P,Q_{\theta_{\rm out}})\le C\min_k(\h^2(P,Q_{\theta_k})+t_k)+C\log(2/\xi)/n_{\rm val}\) with failure probability \(\xi\) (Appendix~\ref{app:validation}).

To prove Theorem~\ref{thm:proper}, fix a comparator \(v\) that nearly minimizes \(\Delta_v+\gamma_H\excess_v\). The first radius is \(r_1=2\ge\sqrt2\ge\h(Q_{a_1},Q_v)\), so the distance requirement in \eqref{eq:main-stage} holds at the first stage. While \eqref{eq:main-stage} holds, a stage returns an anchor within \(r/2\), since \eqref{eq:direction-main} and the regret bound of projected updates, or a union of clean blocks, provide a candidate within \(r/\Gamma\) of \(Q_v\) for a large constant \(\Gamma\). Stage \(j\) costs \(O((dL_H+\gamma_H\iota_j)/r_j^2)\) trajectories and fails with probability \(O(e^{-\iota_j})\); these sum to at most \(\delta/2\). The anchor of the last stage at which \eqref{eq:main-stage} holds has squared Hellinger error \(O(\varepsilon+\Delta_v+\gamma_H\excess_v)\). The final \(\textsc{Select}\) weights anchor \(j\) by \(2^{j-J_\varepsilon}\), whose penalty grows linearly as the index moves back while \(r_j^2\) grows geometrically, so it recovers this error from \(C\log(2/\delta)/\varepsilon\) trajectories without identifying the stage. Summing the stage budgets gives
\[
 O\Bigl(\frac{1}{\varepsilon}\sum_{k\ge0}4^{-k}
             [dL_H+\gamma_H(\log(2/\delta)+k+1)]\Bigr),
\]
the sample bound of Theorem~\ref{thm:proper}. All steps run in polynomial time (Appendices~\ref{app:filtered}--\ref{app:validation}).

\section{Computational lower bound for proper learning}
\label{sec:lower}

The proof is a reduction from Label Cover; Appendix~\ref{app:lower} gives the full construction and verifies the constants.

\subsection{The instance}

A Label Cover instance is a bipartite graph in which each edge \(e=(u,v)\) carries a map \(\pi_e\) from labels at \(u\) to labels at \(v\). For every fixed soundness threshold, it is NP-hard to distinguish instances in which some labeling satisfies every edge from instances in which every labeling satisfies at most a threshold fraction of the edges, with alphabets of constant size \citep[Theorem~5]{bhangale26}. A parameter assigns each vertex a probability vector over its labels, so \(\Theta\) is a product of simplices. A context draws an edge, independent uniform residues \(U_i\in\{0,\ldots,m-1\}\) indexed by the right labels, and a uniform sign \(b\in\{-1,1\}\). With endpoint vectors \(\ell_u\) and \(r_v\), let
\[
 F_b=\sum_s\ell_{u,s}[U_{\pi_e(s)}+b]_m
                         -\sum_i r_{v,i}U_i ,
\]
where \([\cdot]_m\) denotes reduction modulo \(m\). The modular reduction keeps every logit bounded, since all residues lie in \(\{0,\dots,m-1\}\), while a shift by \(b\) remains linear in the soft labels except on wrap-around events, which have probability \(1/m\) for each label. The residues are attached to the right labels, and the left labels read them through \(\pi_e\) (Figure~\ref{fig:labelcover}a). The model uses logit \(F_b/m\) at every token, the target uses logit \(b/m\), and both generate conditionally independent tokens. The logit is linear in the parameter, the features satisfy the norm bounds of Section~\ref{sec:model}, and projection onto \(\Theta\) is a product of simplex projections.

A satisfying labeling yields one-hot vectors with \(F_b=b\) unless the selected residue wraps around modulo \(m\), which happens with probability \(1/m\). Hence \(\OPT\le2/m\) on satisfiable instances.

\subsection{Soundness by rounding soft labels}

\begin{figure}[t]
\centering
\includegraphics{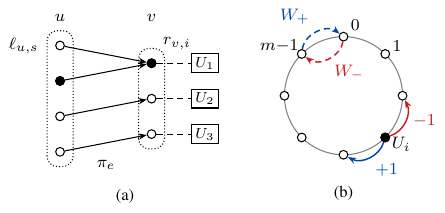}
\vskip -0.05in
\caption{The construction behind Theorem~\ref{thm:lower}. (a) One Label Cover edge \(e=(u,v)\): circles are labels, arrows are \(\pi_e\), filled circles form a satisfying labeling, and \(\ell_{u,s}\), \(r_{v,i}\) are soft labels. Right label \(i\) carries the residue \(U_i\). (b) On \(\mathbb Z_m\), the sign shifts the residue read by the left endpoint by \(+1\) (blue) or \(-1\) (red), both acting on the same \(U_i\). Dashed arcs are the wrap-arounds \(m-1\mapsto0\) and \(0\mapsto m-1\); \(W_\pm\) is \(m\) times the left soft-label mass on right labels whose residue wraps (Section~\ref{sec:lower}).}
\label{fig:labelcover}
\vskip -0.1in
\end{figure}

The rounding argument uses both signs on the same residue table, as in Figure~\ref{fig:labelcover}b. On a fixed edge, let \(\mu_i=\sum_{s:\pi_e(s)=i}\ell_{u,s}\) be the left soft labels pushed forward to the right labels, and write
\[
 S=\textstyle\sum_i(\mu_i-r_{v,i})U_i,\quad
 W_\pm=m\textstyle\sum_i \mu_i\ind_{\{U_i=u_\pm\}},
\]
with \(u_+=m-1\) and \(u_-=0\).
Modular arithmetic gives \(F_+-1=S-W_+\) and \(F_-+1=S+W_-\). If both errors have magnitude at most \(a=1/16\) on a table, then \(W_++W_-\le2a\) and \(|S|\le a\), since \(W_\pm\ge0\). A single sign would constrain only a difference such as \(S-W_+\); the two signs on the same table bound the wrap mass and the moment \(S\) separately.

Suppose that the event \(|F_b-b|>a\) had probability less than \(\beta=1/200\) over the contexts. Then on at least half of the edges the conditional failure probability is at most \(1/100\), and on each such edge both signs fit on the same table with probability at least \(0.98\). The wrap constraint forces the left soft labels to concentrate. Indeed, \(W=W_++W_-\) has mean two for every choice of soft labels and second moment at most \(4+2m\|\mu\|_2^2\), but it is at most \(2a\le1\) with probability at least \(0.98\); its mass above one must then come from a rare event, and Cauchy--Schwarz on \(\{W>1\}\) forces \(m\|\mu\|_2^2\ge23\). The moment constraint forces the two endpoints to agree. The variance of \(S\) is \((m^2-1)\|\mu-r_v\|_2^2/12\) and \(\E S^4\le3(\E S^2)^2\), so if \(\E S^2>2a^2\), the Paley--Zygmund inequality would give \(\Prob(|S|>a)\ge1/12\), which is incompatible with both signs fitting. Hence \(\|\mu-r_v\|_2\le6a/m\), and
\[
      \mu^\top r_v\ge\|\mu\|_2^2-\|\mu\|_2\|\mu-r_v\|_2
              \ge (23-6a)/m>1/m .
\]
Concentration and agreement together make randomized rounding effective: independent rounding of each vertex according to its soft labels satisfies the edge with probability \(\mu^\top r_v\), and would therefore satisfy more than a \(1/(2m)\) fraction of the edges in expectation, which contradicts soundness \(1/(4m)\). On instances of low value, every feasible parameter has \(|F_b-b|>a\) on a fraction at least \(\beta\) of the contexts.

\subsection{Detecting the error and the proof of Theorem~\ref{thm:lower}}

A logit error above \(a/m\) moves the Bernoulli mean by more than \(a/(6m)\), as \(\sigma'>1/6\) on \([-1,1]\), while the frequency of ones among \(H=C_0m^2\) tokens concentrates within \(O(1/(m\sqrt{C_0}))\) of its mean. At this horizon a single trajectory therefore detects the error with constant statistical power. We reject a trajectory when its frequency of ones is more than \(a/(12m)\) away from the known target mean \(\sigma(b/m)\). For \(C_0\) large, the target rejects with probability at most \(\beta/16\), while on an instance of low value every feasible parameter, and hence every context-independent mixture, rejects with probability at least \(\beta(1-\beta/16)\).

Now suppose that an algorithm as in Theorem~\ref{thm:lower} exists, with \(\varepsilon_0\) and \(c\sqrt{C_0}\) small compared with \(\beta^2\), and fix \(m\ge2\). On a satisfiable instance \(\sqrt H\,\OPT\le2\sqrt{C_0}\), so with probability \(9/10\) the output has squared Hellinger error below \((\beta/8)^2\) and, since total variation is at most \(\h\), rejects with probability below \(\beta/4\). We simulate the data and both oracles from the graph and the public mean, estimate the rejection probability of the output from a constant number of rollouts, and accept when the estimate is below \(\beta/2\). This accepts satisfiable instances and rejects instances of low value, each with probability at least \(2/3\). For fixed \(m\) the alphabets and \(H\) are constants and \(d\) grows with the graph, so the procedure decides an NP-hard gap problem in randomized polynomial time, contradicting \(\mathrm{NP}\not\subseteq\mathrm{BPP}\); the constants do not depend on \(m\).

\section{Discussion and open problems}
\label{sec:discussion}

\emph{Dimension-dependent optimality.} What is the optimal factor as a joint function of \(d\) and \(H\)? Theorem~\ref{thm:proper} gives \(O(\min\{d,\sqrt H\})\), while Theorem~\ref{thm:lower} fixes \(H\) and lets \(d\) grow. A lower bound for \(d\lesssim\sqrt H\), or a learner whose factor there does not grow with \(H\), would complete the picture.

\emph{Beyond proper learning.} Can efficient improper learners beat \(O(\sqrt H)\)? An unrestricted generator fits our hard targets by reading \(b\), and the best improper lower bound \citep{rohatgi25} excludes only factors \(e^{\log^{1-c}\max\{d,H\}}\).
\label{sec:main-end}

\clearpage
\section*{Impact Statement}
This paper presents theoretical work on learning autoregressive sequence models from offline data under model misspecification. It characterizes how the approximation error of a learned model can grow with the sequence length, and what computationally efficient learners can achieve. We do not foresee specific societal consequences of this work that need to be highlighted here.

\section*{AI Disclosure Statement}
Generative AI assisted with developing the algorithms, deriving and checking mathematical arguments, reviewing related work, drafting and revising this manuscript, and drawing its two figures in TikZ. OpenAI Codex was used for earlier drafts, and Anthropic Claude was used to restructure and rewrite the present version and to audit its proofs. The draft was checked by tracing the assumptions and inequalities in its proofs, by enumerating selected finite adaptive binary trees, and by numerical tests of selected second-moment and rounding calculations. The authors take full responsibility for the content of the paper.

\bibliographystyle{icml2027}
\bibliography{references}

\clearpage
\appendix
\onecolumn
\section{Proof of Theorem~\ref{thm:proper}}
\label{app:proper}

Throughout this appendix, \(L_H=\log(2+H)\) and \(w_H=\sqrt H\), and \(C,c\) denote absolute constants whose values may change between occurrences; subscripts indicate dependence on fixed tuning constants. We fix a comparator \(v\in\Theta\) and analyze one stage of Algorithm~\ref{alg:proper}. Conditional on the preceding stages, the current anchor \(a\), the radius \(0<r\le2\) and the confidence parameter \(\iota\ge1\) are fixed, and all samples used by the stage are fresh. Appendices~\ref{app:geometry}--\ref{app:small} show that the stage returns an anchor within Hellinger distance \(r/2\) of \(Q_v\) whenever \eqref{eq:main-stage} holds, and Appendix~\ref{app:validation} chains the stages and proves the theorem.

\subsection{Local geometry}
\label{app:geometry}

The analysis compares parameters through the quadratic forms of the design matrix \(A\) under different laws. For a law \(S\) on trajectories write \(M_S=\E_S A\) and \(\|u\|_M=\sqrt{u^\top M u}\), and put
\[
 \|u\|_*=\mathop{\rm ess\,sup}_{x\sim\cD}\ \max_{y_{1:H}}\sqrt{u^\top A(x,y_{1:H})u},
 \qquad
 {\cal N}_r(u)=\|u\|_{M_P}+r\|u\|_* .
\]
The norm \(\|\cdot\|_*\) is the worst-case energy, and \({\cal N}_r\) mixes it with the population energy at the scale of the stage. Two laws that are close in Hellinger distance have similar population energies. Indeed, for laws \(S,S'\) with context marginal \(\cD\), \(\|u\|_{M_S}\) is the \(L^2\) norm of \((u^\top Au)^{1/2}\sqrt{s}\), and the triangle inequality in \(L^2\) gives
\begin{equation}
 \big|\|u\|_{M_S}-\|u\|_{M_{S'}}\big|
                           \le \h(S,S')\|u\|_* .
 \label{eq:design-transfer}
\end{equation}
Since \(|\ip{\theta}{z_h}|\le2\) for every feasible \(\theta\), feasible differences satisfy \(\|w-v\|_*^2\le16H\). Likelihoods and all design quadratic forms are invariant along the common null space of the design matrices, so we factor this null space out whenever we use a norm ball and its volume, while computations stay in the original coordinates.

Energy and Hellinger distance control each other in both directions. Since \(\min\{1,V\}\ge V/(16H)\) whenever \(0\le V\le16H\), the capped energy bound \eqref{eq:energy-capped} of Section~\ref{sec:inputs} implies \(\E_{Q_v}(w-v)^\top A(w-v)\le16CH\,\h^2(Q_v,Q_w)\), that is,
\begin{equation}
 \|w-v\|_{M_{Q_v}}\le C w_H\,\h(Q_v,Q_w)
                         \qquad (v,w\in\Theta).
 \label{eq:energy}
\end{equation}
Conversely, the log-partition function of a Bernoulli logit has second derivative at most \(1/4\), so each token contributes at most \(\ip{u}{z_h}^2/8\) to the Kullback--Leibler divergence, and the chain rule gives
\begin{equation}
  \h^2(Q_w,Q_{w+u})\le\KL(Q_w\|Q_{w+u})
       \le \|u\|_{M_{Q_w}}^2/8 .
 \label{eq:smooth-kl}
\end{equation}
This upper bound does not use feasibility, and it holds for every translate \(w+u\), with \(Q_{w+u}\) defined by the same logistic formula.

The ellipsoid \(K_r\) is built from the pilot average \(\widehat M\), so we need \(\widehat M\) to approximate \(M_P\) in the norm \({\cal N}_r\). For any small fixed \(\vartheta>0\), a pilot of \(C_\vartheta(d+\iota)/r^2\) fresh trajectories from \(P\) ensures, with failure probability at most \(e^{-\iota}\),
\begin{equation}
  |u^\top(\widehat M-M_P)u|\le \vartheta\,{\cal N}_r(u)^2
                  \quad\text{for all }u .
 \label{eq:design-pilot}
\end{equation}
To see this, note that on the \({\cal N}_r\)-unit sphere each quadratic observation \(u^\top A(o)u\) is at most \(\|u\|_*^2\le r^{-2}\) and has mean \(\|u\|_{M_P}^2\le1\). Bernstein's inequality therefore gives a tail bound \(2\exp(-c_\vartheta nr^2)\) for a fixed direction, using \(r\le2\). A \(1/32\)-net of the unit ball has at most \(65^d\) points, and by polarization the supremum of a symmetric bilinear form over the ball is at most twice the supremum of its quadratic form, so control on the net extends to the whole sphere. The same argument shows \(\widehat A\preceq C{\cal N}_r^2\), as quadratic forms, for an independent fitting batch from \(P\). Under \eqref{eq:main-stage}, \(\h(P,Q_v)\le\sqrt{c_*}\,r\), so \eqref{eq:design-transfer} gives \(\|u\|_{M_{Q_v}}\le\|u\|_{M_P}+\sqrt{c_*}\,r\|u\|_*\le C{\cal N}_r(u)\), and the bound on the fitting design average also holds for fresh samples from \(Q_v\).

We can now locate the comparator and all candidates. Let \(K=K_r\) be as in \eqref{eq:algorithm-pilot}. Under \eqref{eq:main-stage}, \eqref{eq:energy} gives \(\|a-v\|_{M_{Q_v}}\le Cw_Hr\), \eqref{eq:design-transfer} moves this to \(M_P\) at a cost \(\h(P,Q_v)\|a-v\|_*\le4\sqrt{c_*}\,w_Hr\), and \eqref{eq:design-pilot} moves it to \(\widehat M\) at a cost \(\sqrt\vartheta\,{\cal N}_r(a-v)\le Cw_Hr\). Hence \(\|a-v\|_{\widehat M}\le C_Kw_Hr\) for \(C_K\) large, that is, \(v\in K\). In the other direction, every \(w\in K\) satisfies \(\|w-a\|_{M_P}\le\|w-a\|_{\widehat M}+\sqrt{\vartheta}{\cal N}_r(w-a)\) and \(\|w-a\|_*\le4w_H\); absorbing the small multiple of \({\cal N}_r\) gives \({\cal N}_r(w-a)\le C w_H r\), and in particular \({\cal N}_r(w-v)\le Cw_H r\). Applying \eqref{eq:smooth-kl} from \(Q_v\) yields the crude bound \(\h(Q_v,Q_w)\le Cw_H r\), and then \eqref{eq:design-transfer} gives
\begin{equation}
      \|u\|_{M_{Q_w}}\le Cw_H\,{\cal N}_r(u)
          \qquad\text{for all }u\text{ and }w\in K .
 \label{eq:local-model}
\end{equation}

\subsection{Local uniform likelihood comparison}
\label{app:local}

This subsection proves the local likelihood bound (L) of Section~\ref{sec:retention}. We use the event \(\cI\), the distance \(h_w\) and the capped log-likelihood ratio \(Y_w\) of Section~\ref{sec:score-likelihood}. The fixed-parameter bound \eqref{eq:likelihood-moment} also holds, by the same affinity calculation, with \(\cI\) the whole space and \(P=Q_v\).

A union bound over a grid of candidates at precision \(\varepsilon\) would cost a factor \(\log(1/\varepsilon)\) in the sample size. We avoid it with the following entropy inequality. If \(\E_S\exp f_\theta\le1\) for each fixed \(\theta\) and the prior \(\pi_0\) is independent of the fitting sample, then with probability at least \(1-e^{-\iota}\),
\[
 \int \sum_{i=1}^n f_\theta(o_i)\,d\mu(\theta)
                       \le \KL(\mu\|\pi_0)+\iota
\]
simultaneously for every posterior \(\mu\) absolutely continuous with respect to \(\pi_0\). This follows by integrating the product exponential over \(\pi_0\), applying Markov's inequality, and applying Jensen's inequality to the density \(d\mu/d\pi_0\).

We use as prior the uniform law on the \({\cal N}_r\)-ball of radius \(C w_H r+r\) centered at the anchor, and for \(w\in K\) we use as posterior the uniform law on the \({\cal N}_r\)-ball of radius \(\varsigma r/w_H\) centered at \(w\). The two radii differ by a factor \(O(H/\varsigma)\), so the relative entropy is \(O_\varsigma(d L_H)\). By \eqref{eq:local-model} and \eqref{eq:smooth-kl}, every translate \(w+u\) in the small ball is within Hellinger distance \(C\varsigma r\) of \(Q_w\), so the triangle inequality shows that the posterior mean of \(h_{w+u}^2\) is at least \(h_w^2/2-C\varsigma^2r^2\).

It remains to compare the posterior average of the capped log ratio with its value at \(w\). The negative log likelihood has Hessian at most \(A(o)/4\), so pointwise the capped log ratio at \(w+u\) is at most its value at \(w\), plus a linear form in \(u\), plus \(u^\top A(o)u/8\); the linear form is the gradient of the uncapped log ratio when the cap is inactive and zero when it is active. Multiplying by \(\ind_\cI\) and averaging over the small ball, the linear term vanishes by symmetry, and on the event \(\widehat A\preceq C{\cal N}_r^2\) the remainder contributes at most \(C\varsigma^2 r^2\) to the fitting average. Applying the entropy inequality to \eqref{eq:likelihood-moment} and combining these estimates shows that, simultaneously for all \(w\in K\),
\begin{equation}
 P_nY_w\ \ge\ c_1 h_w^2-C\Delta_v-\zeta r^2
                      -C_\zeta(d L_H+\iota)/n ,
 \label{eq:local-likelihood}
\end{equation}
where \(\zeta>0\) is any prescribed constant and \(\varsigma\) is chosen small enough in terms of \(\zeta\). The same argument applies to an independent fitting batch from \(Q_v\), with \(\cI\) the whole space and \(\Delta_v=0\). Only \eqref{eq:smooth-kl} and \eqref{eq:likelihood-moment} are applied at infeasible translates, and both hold there.

\subsection{Score moments, leverage and the second-moment witness}
\label{app:witness}

This subsection proves the witness bound (W) of Section~\ref{sec:retention}. The stage uses \(n\) pilot trajectories and \(n\) fitting trajectories with the ridge of \eqref{eq:algorithm-pilot}; on \(K-K\) the ridge quadratic form is at most \(\zeta r^2\), by the ellipsoid constraint and the radius bound \(B\). The witness must show that the scores of the comparator satisfy the second-moment constraint after a reweighting whose logarithmic cost is small. We first control the score in a single direction, then pass to all directions at once, and finally convert the result into weights by convex duality.

\paragraph{A scalar score test.}
Fix a direction \(t\) and let \(X_t=\ip{t}{G_v}\) and \(V_t=t^\top A t\). Under \(Q_v\), \(X_t\) is a sum of martingale differences whose increments have conditional ranges \(|\ip{t}{z_h}|\), also when the contrasts depend on earlier tokens, so iterating conditional Hoeffding bounds along the trajectory gives \(\E_{Q_v}\exp(\pm X_t-V_t/8)\le1\). Averaging over the two signs and using \(\cosh(x)\ge1+x^2/2\) yields a moment bound for
\[
 T_t=\max\{-4,\log(1+X_t^2/2)-V_t/8\}.
\]
Since \(e^{\max(-4,s)/2}\le(1+e^s)/2\) for every real \(s\), it follows that \(\E_{Q_v} e^{T_t/2}\le1\). To transfer this bound from \(Q_v\) to \(P\), put \(D_t=e^{T_t/4}-1\); expanding \(e^{T_t/2}=(1+D_t)^2\) gives \(\E_{Q_v}(D_t+D_t^2/2)\le0\). As in Section~\ref{sec:score-likelihood}, changing measure on \(\cI\) costs at most \(3\sqrt{\Delta_v\,\E_{Q_v}D_t^2}\), and dropping the complement costs at most \(\Delta_v\) because \(D_t\ge-1\). The quadratic term absorbs the square-root cost, and we obtain
\begin{equation}
       \E_P\exp(\ind_\cI T_t/4)\le e^{C\Delta_v}.
 \label{eq:score-moment}
\end{equation}
Because of the lower truncation at \(-4\), omitting one trajectory changes the sum of these statistics by at most four.

\paragraph{Leverage of fresh trajectories.}
To pass from one direction to all directions, the Gaussian comparison below needs the fitting design matrices to be small relative to the pilot. The population design may be poorly conditioned, and no eigenvalue lower bound is assumed. Instead, a trace identity bounds the number of fitting trajectories whose design matrix has large leverage relative to the pilot. Below, a row is the design matrix of one trajectory.

\begin{lemma}[Leverage of fresh trajectories]\label{lem:leverage}
Let \(A_i\) be independent and identically distributed positive semidefinite \(d\times d\) matrices, and fix a deterministic \(R_0\succ0\). On a pilot of size \(n\ge2\) set \(D_0=R_0+\sum_{\rm pilot} A_i\). On an independent fitting batch of the same size,
\[
   \sum_{\rm fit}\min\{1,\tr(D_0^{-1}A_i)\}\le C(d+\iota)
\]
except on an event of probability at most \(C e^{-\iota}\).
\end{lemma}
\begin{proof}
For a list \(S\) of matrices, put \(b(A,S)=\min(1,\tr((R_0+\sum_{j\in S}A_j)^{-1}A))\), which decreases when \(S\) is extended. For an element \(A_i\) of the list, let \(\bar M=R_0+\sum_{j\ne i}A_j\). If \(\lambda_j\) are the eigenvalues of \(\bar M^{-1/2}A_i\bar M^{-1/2}\), then
\[
 \tr((\bar M+A_i)^{-1}A_i)
   =\sum_j\frac{\lambda_j}{1+\lambda_j}
   \ge\frac{\sum_j\lambda_j}{1+\sum_j\lambda_j}
    \ge \tfrac12 b(A_i,S\setminus\{i\}),
\]
and the left-hand sides sum to at most \(d\) over the list, since they are the leverages \(\tr(D^{-1}A_i)\) of the full sum \(D=R_0+\sum_jA_j\).

On the pilot, put \(B_i=b(A_i,[i-1])\) for \(i=\lfloor n/2\rfloor+1,\ldots,n\). Condition on the unordered list and reveal its order backwards. At step \(i\) the last remaining row is uniform among \(i\) rows, so by the identity above the conditional mean of \(B_i\) is at most \(2d/i\). Since \(0\le B_i\le1\), the inequality \(e^x\le1+(\mathrm e-1)x\) on \([0,1]\) and iterated conditional moments give \(\sum_i B_i\le C(d+\iota)\) with failure probability at most \(e^{-\iota}\). Now consider the forward order and put \(\mu_i=\E[B_i\mid A_{<i}]\). The inequality \(e^{-x}\le1-(1-e^{-1})x\) on \([0,1]\) gives \((1-e^{-1})\sum_i\mu_i\le\sum_i B_i+\iota\) with failure probability at most \(e^{-\iota}\). Since \(b\) decreases as the list grows, the conditional mean of the leverage of a fresh matrix with respect to the whole pilot is at most every \(\mu_i\), hence at most \(C(d+\iota)/n\) on the joint event. Applying the first exponential bound to the independent fitting batch, conditional on the pilot, completes the proof.
\end{proof}

We apply the lemma with \(R_0=n\lambda I_d\), so that \(D_0=n(\widehat M+\lambda I_d)\). At most \(C(d+\iota)\) fitting rows then have leverage above one, and the remaining rows have total leverage at most \(C(d+\iota)\). Let \(\widehat A_{\rm lo}\) be the sum of the design matrices of fitting rows with leverage at most one, divided by the full fitting size \(n\), and let \(\cI_0\) be the set of these trajectories that lie in \(\cI\).

\paragraph{A Gaussian comparison.}
To combine the scalar bounds \eqref{eq:score-moment} over all directions, we integrate the product of the tests over a Gaussian prior on the direction \(t\), namely \(N(0,(nM_0)^{-1})\), which is fixed conditional on the pilot. Markov's inequality and the entropy variational inequality give an empirical bound that holds simultaneously for all Gaussian posteriors, at an additional cost of \(\iota\) plus the relative entropy to the prior. For a positive semidefinite \(\Lambda\), we use the posterior \(N(0,2\Lambda+\Sigma_{\rm lo})\) with
\(\Sigma_{\rm lo}=[n(M_0+\widehat A_{\rm lo})]^{-1}\).
Under this posterior, \(X_t\) on a fixed row is a centered Gaussian with variance at least \(2G_v^\top\Lambda G_v\); its square exceeds this variance with constant probability, so the posterior mean of \(\log(1+X_t^2/2)\) is at least a constant multiple of \(\log(1+G_v^\top\Lambda G_v)\). The posterior mean of the quadratic term \(V_t\) on \(\cI_0\) is at most \(2\tr(\Lambda\widehat A)+d/n\), because \(\tr(\widehat A_{\rm lo}\Sigma_{\rm lo})\le d/n\). By the formula for the relative entropy between Gaussians, the entropy cost is at most
\[
 n\tr(\Lambda M_0)
     +\tfrac12\log\det(I_d+M_0^{-1/2}\widehat A_{\rm lo}M_0^{-1/2});
\]
here the baseline \(\Sigma_{\rm lo}\) contributes at most \(d\) to the trace, and adding \(2\Lambda\) only increases the determinant of the posterior covariance. The leverage bound now enters. On the event of Lemma~\ref{lem:leverage}, the normalized design matrix in the determinant has rank at most \(d\) and trace \(O((1+H)(d+\iota))\), so the arithmetic--geometric mean inequality bounds the log determinant by \(O(dL_H+\iota)\). Dividing by \(n\) and charging the at most \(C(d+\iota)\) omitted rows gives
\begin{equation}
 P_n\ind_{\cI_0}\log(1+G_v^\top\Lambda G_v)
 \le C\tr(\Lambda(\widehat A+M_0))
           +C_{\lambda_*}\{\Delta_v+(dL_H+\iota)/n\}.
\label{eq:gaussian-cov}
\end{equation}

\paragraph{The witness.}
Inequality \eqref{eq:gaussian-cov} holds for every positive semidefinite \(\Lambda\), and convex duality turns it into weights. Precisely, there are weights \(0<y_i\le1\) on \(\cI_0\) such that
\begin{equation}
 \begin{split}
 P_n\ind_{\cI_0}(-\log y_i)&\le C_{\lambda_*}\{\Delta_v+(dL_H+\iota)/n\},\\
 P_n\ind_{\cI_0} y_i^2G_vG_v^\top&\preceq C(\widehat A+M_0).
 \end{split}
 \label{eq:witness}
\end{equation}
To see this, minimize the first left-hand side subject to the second constraint. Because of the ridge, sufficiently small strictly positive weights satisfy the matrix constraint and the box constraints strictly, so strong duality holds and the minimum is attained. The Lagrange dual has a positive semidefinite multiplier \(\Lambda\), and each retained row contributes \(\min_{0<y\le1}[-\log y+(G_v^\top\Lambda G_v)y^2]\). With \(\upsilon=G_v^\top\Lambda G_v\ge0\), this minimum equals \(\upsilon\) for \(\upsilon\le1/2\) and \((1+\log(2\upsilon))/2\) otherwise, and in both cases it is at most \(2\log(1+\upsilon)\). With a large enough design multiplier, the linear charge \(\tr(\Lambda(\widehat A+M_0))\) of the constraint absorbs the trace term in \eqref{eq:gaussian-cov}, and \eqref{eq:witness} follows; when \(\cI_0\) is empty the claim holds with empty sums. The logarithm in the dual cost is the reason for the logarithmic objective in \eqref{eq:program}. The weights \(y_i\) are used only in the analysis; the program \eqref{eq:program} uses all fitting rows and depends only on the scores at \(w\), the observed design matrices and the pilot ridge.

\subsection{Filtered gradient and polynomial implementation}
\label{app:filtered}

Let \(\tau=1/w_H\), \(a_0=\kappa/24\) and \(L_\tau=\log(1+1/\tau)\). We solve \eqref{eq:program} with a sufficiently large design multiplier \(C_A\), objective tolerance \(\zeta r^2\), and a second copy of \(M_0\) as feasibility slack. The matrix constraint is handled by a penalty. With \(\mathcal V(\omega)=P_n\omega_i^2G_wG_w^\top-C_A(\widehat A+M_0)\), we minimize the negative log objective plus
\[
 \frac{1+L_\tau}{C_A\lambda_{\min}(M_0)}
                      [\lambda_{\max}(\mathcal V(\omega))]_+ ,
\]
where an inverse-polynomial lower bound on the smallest eigenvalue may replace \(\lambda_{\min}(M_0)\). The penalized objective is convex, and a top eigenvector of \(\mathcal V(\omega)\) gives a subgradient of the active quadratic. A feasible comparison weight pays no penalty and has negative log objective at most \(L_\tau\), so an approximate minimizer to tolerance at most \(\zeta r^2\le1\) attains the required objective and violates the constraint by at most the slack. Projected subgradient descent on the unit box needs polynomially many steps, because the box has diameter \(\sqrt n\), the ridge is at least \(\lambda_*\lambda/(1+H)\), the scores have norm at most \(2BH\), and the derivative of the logarithm is at most \(1/\tau\). An approximate eigenvector with inverse-polynomial Rayleigh error gives a subgradient with a correspondingly small error, as follows by comparing its quadratic form with the supremum that defines the top eigenvalue.

With the inputs \eqref{eq:witness} and \eqref{eq:local-likelihood}, the leverage event of Lemma~\ref{lem:leverage} and the binomial bound on \(\exfrac\), the deterministic argument of Section~\ref{sec:retention} yields the directional bound \eqref{eq:direction-main} uniformly on \(K\). Each filtered gradient has norm at most \(6BH\), because the weights are at most one and \(\tau\le1\).

The directional bound becomes contraction through projected updates. If \(w_{t+1}\) is the projection of \(w_t-\eta g(w_t)\) onto \(K\), then, since \(v\in K\),
\[
 2\eta\ip{w_t-v}{g(w_t)}
   \le\|w_t-v\|_2^2-\|w_{t+1}-v\|_2^2+\eta^2(6BH)^2 .
\]
Summing over \(T\) steps with \(\eta=2B/(6BH\sqrt T)=1/(3H\sqrt T)\) and using \(\|w_1-v\|_2\le2B\) gives \(\sum_{t=1}^T\ip{w_t-v}{g(w_t)}/T\le12 B^2H/\sqrt T\), which a polynomial \(T\) makes smaller than any fixed multiple of \(r^2\). Because \eqref{eq:direction-main} holds uniformly on \(K\), all iterations can reuse the same fitting batch.

Exact projections are not needed. Let \(U(w)=\|w-a\|_{\widehat M}\) and \(R_r=C_K w_H r\), and minimize the projection objective over \(\Theta\) plus a large polynomial multiple of \((U(w)-R_r)_+\). The radial repair \(w\mapsto a+\min(1,R_r/U(w))(w-a)\) stays in \(\Theta\) by convexity and moves the point by at most \(2B(U(w)-R_r)_+/R_r\). Since the squared-distance objective is polynomially Lipschitz on the bounded set, the penalty pays for this movement, and strong convexity places the repaired point close to the exact projection. This convex penalized problem can be solved to inverse-polynomial precision by projection onto \(\Theta\) and subgradient steps. The same repair and Lipschitz estimate allow minimizing the convex mean negative log likelihood over \(K\) to any inverse-polynomial tolerance, as required in Appendix~\ref{app:small}. Errors in the weighted or unweighted score averages and in the projections enter the squared-distance recursion linearly in their magnitudes, up to polynomial factors, so their total can be made arbitrarily small relative to \(r^2\). Every radius used before the final validation is at least a constant multiple of \(\sqrt\varepsilon\), so these tolerances have polynomial cost.

We can now complete a stage in the branch \(w_H<d\). Fix a sufficiently large numerical contraction constant \(\Gamma\), take \(\zeta\) and \(c_*\) small enough, and let \(n\ge C(dL_H+w_H \iota)/r^2\). Under \eqref{eq:main-stage}, the subtracted terms in \eqref{eq:direction-main} are then at most a small multiple of \(r^2\), and so is the average regret. Averaging \eqref{eq:direction-main} over the iterates therefore bounds the average of \(h_{w_t}^2\) by a small multiple of \(r^2\), and Markov's inequality shows that at least nine tenths of the iterates satisfy \(\h(Q_{w_t},Q_v)\le r/\Gamma\), after increasing numerical constants. Drawing \(C\iota\) indices uniformly at random yields a candidate this close with failure probability at most \(e^{-\iota}\). A fresh validation batch of \(C_\Gamma(1+\iota)/r^2\) trajectories then gives an anchor within \(r/2\) of \(Q_v\), by \eqref{eq:validate} below and \(\h(P,Q_v)^2\le c_*r^2\).

\subsection{Small-dimension branch}
\label{app:small}

When \(d\le w_H\), the improvement condition in \eqref{eq:main-stage} reads \(\Delta_v+d\excess_v\le c_* r^2\), and the stage uses likelihood fits on blocks of data instead of the filtered gradient. It uses the same independent pilot and the same region \(K\).

We first consider clean data. Under \(N\) independent trajectories from \(Q_v\), an approximate likelihood fit over \(K\) lies within Hellinger distance \(r/\Gamma\) of \(Q_v\) with failure probability at most \(\exp(-c_\Gamma N r^2)\), provided \(N\ge C_\Gamma d L_H/r^2\). Indeed, \eqref{eq:local-likelihood} with \(P=Q_v\), applied with \(\zeta\) small and \(\iota=c_\Gamma N r^2\) and with a large sample constant, bounds the capped empirical likelihood ratio from below by a fixed positive multiple of \(r^2/\Gamma^2\) at every candidate farther than \(r/\Gamma\) from \(Q_v\). The uncapped log ratio is at least the capped one, and \(v\) is feasible, so a likelihood fit whose mean log likelihood is within a small \(c'_\Gamma r^2\) of the optimum cannot be such a distant candidate.

The target differs from \(Q_v\) in two ways, a bounded density ratio and an excess component, and the two are handled separately. Write \(P=(1-\excess_v)P_0+\excess_v P_1\), where \(P_0\) is the normalization of the subdensity \(\min(p,2q_v)\) and \(P_1\) the normalization of \((p-2q_v)_+\); in a stage satisfying \eqref{eq:main-stage}, \(\excess_v\le1/2\). Then \(dP_0/dQ_v\le4\), and the affinity bound \(\int\sqrt{p_0p}\ge\sqrt{1-\excess_v}\) gives
\(\h^2(P_0,P)\le2(1-\sqrt{1-\excess_v})\le2\excess_v\). Since the density ratio is at most four, \(\chi^2(P_0,Q_v)\le9\,\h^2(P_0,Q_v)\), and the triangle inequality with \(\excess_v\le2\Delta_v\) gives \(\chi^2(P_0,Q_v)\le C\Delta_v\). The Cauchy--Schwarz inequality applied to the product likelihood ratio then gives, for any event \(E\),
\(P_0^{\otimes N}(E)\le e^{C N\Delta_v}\sqrt{Q_v^{\otimes N}(E)}\).
For \(c_*\) small enough, the guarantee of the likelihood fit therefore transfers to \(P_0^{\otimes N}\), with failure probability at most \(\exp(-c''_\Gamma N r^2)\). This transfer compares joint laws, so it remains valid even though \(P_0\) may have a different context marginal.

The excess component is avoided by enumerating unions of blocks. Set \(J=\lceil C_J L_H\rceil\) for a fixed constant \(C_J\) and \(n_0=\lceil C_\Gamma d/r^2\rceil\). A pool consists of \(4J\) independent blocks of \(n_0\) trajectories, and the learner fits the likelihood on each union of \(J\) blocks. In the latent decomposition above, each trajectory comes from \(P_0\) or from \(P_1\), and we call a block clean if all its trajectories come from \(P_0\); the learner never observes this decomposition. A block is clean with probability \((1-\excess_v)^{n_0}\ge3/4\), by the stage condition with a small enough \(c_*\), so by a binomial bound at least \(J\) blocks of a pool are clean, except with probability \(\exp(-cJ)\). Conditional on the latent marks, the first \(J\) clean blocks have law \(P_0^{\otimes n_0J}\), and their fit is close to \(Q_v\) except with probability \(\exp(-c''_\Gamma n_0Jr^2)\). A pool therefore contains a close candidate, except with probability \(\exp(-c_3 J)\) for \(C_J\) large. Using \(C(1+\lceil \iota/J\rceil)\) independent pools with the same pilot, at least one candidate is close except with probability \(e^{-\iota}\), conditional on a successful pilot. Each pool requires \(\binom{4J}{J}\le 2^{4J}\) fits, a number polynomial in \(H\). The pools cost \(O(d(L_H+\iota)/r^2)\) trajectories, and validation costs \(O((J+\iota)/r^2)\) trajectories because the logarithm of the number of candidates is \(O(J+\log(2+\iota))\). Independent validation then yields the contraction.

\subsection{Validation and the full schedule}
\label{app:validation}

The same selection rule is used at every stage and for the final list. Let \(\theta_1,\dots,\theta_K\) be candidate parameters, independent of \(n_{\rm val}\) validation trajectories from \(P\), and write \(Q_k=Q_{\theta_k}\) with density \(q_k\). The comparison statistic
\[
 T_{kl}(o)=\frac{\sqrt{q_l(o)}-\sqrt{q_k(o)}}{\sqrt{q_l(o)}+\sqrt{q_k(o)}} ,
\]
with value zero when the denominator vanishes, is bounded by one in absolute value and antisymmetric, \(T_{kl}=-T_{lk}\). Write \(h_k=\h(P,Q_k)\) and \(h_{kl}=\h(Q_k,Q_l)\). Under the model law \(Q_k\), the statistic favors \(Q_k\): with \(f=\sqrt{q_k}\) and \(f'=\sqrt{q_l}\), the identity
\[
       f^2\frac{f'-f}{f+f'}
        =\frac{f'^2-f^2}{4}-\frac{(f'-f)^2}{2}
                                   +\frac{(f'-f)^3}{4(f+f')}
\]
gives \(\E_{Q_k} T_{kl}\le-h_{kl}^2/4\) and \(\E_{Q_k}T_{kl}^2\le h_{kl}^2\). Changing measure to \(P\), we write \(p-q_k=(\sqrt p-\sqrt{q_k})^2+2\sqrt{q_k}(\sqrt p-\sqrt{q_k})\) and apply \(|T_{kl}|\le1\) and the Cauchy--Schwarz inequality, which gives
\[
 \E_P T_{kl}\le-h_{kl}^2/4+2 h_k h_{kl}+h_k^2
                    \le 10 h_k^2-h_l^2/16,
 \qquad
 \E_P T_{kl}^2\le 2h_{kl}^2+2h_k^2 ;
\]
the middle inequality follows from Young's inequality and the triangle inequality \(h_l\le h_k+h_{kl}\). Allocating failure probability \(\xi\pi_k\pi_l\) to the comparison of \(k\) and \(l\), Bernstein's inequality and a union bound give, with failure probability at most \(\xi\),
\[
 \overline T_{kl}\le C h_k^2-h_l^2/32+t_k+t_l+t_\delta
          \quad\text{for all }k,l .
\]
Here \(\overline T_{kl}\) is the validation average, \(t_k=C\log(2/\pi_k)/n_{\rm val}\), \(t_\delta=C\log(2/\xi)/n_{\rm val}\), the \(\pi_k\) are positive weights summing to one, and the variance terms have been absorbed into the squared distances. The rule selects the index \({\rm out}\) that minimizes \(t_k+\max_l(\overline T_{kl}-t_l)\), and it satisfies
\begin{equation}
   \h^2(P,Q_{\rm out})\le C(h_k^2+t_k+t_\delta)
                        \qquad\text{for every }k.
 \label{eq:validate}
\end{equation}
Indeed, the criterion at \(k\) is at most \(Ch_k^2+2t_k+t_\delta\), while comparing against \(k\) and using antisymmetry shows that the criterion at \({\rm out}\) is at least \(h_{\rm out}^2/32-Ch_k^2-2t_k-t_\delta\); minimality of \({\rm out}\) gives \eqref{eq:validate}. Evaluating the rule requires at most one likelihood computation per trajectory and candidate, followed by explicit pairwise operations. All model laws have positive token probabilities, and the computation is stable in log likelihood. Uniform weights \(\pi_k\) give the stage validation bounds.

It remains to chain the stages. The schedule uses \(r_j=2^{2-j}\), so that \(r_1=2\), and \(J_\varepsilon\) is the first index with \(r_j\le c_\varepsilon\sqrt\varepsilon\), where \(c_\varepsilon>0\) is a small universal constant. The initial anchor \(a_1\) is an arbitrary feasible parameter; since \(\h^2\le2\) for any two laws, \(\h(Q_{a_1},Q_v)\le\sqrt2<r_1\), so the distance requirement in \eqref{eq:main-stage} holds at the first stage. Stage \(j<J_\varepsilon\) uses fresh batches and the confidence parameter \(\iota_j=\log(C/\delta)+C(J_\varepsilon-j+1)\), and the initial anchor and every proposed anchor are retained. Summing \(C e^{-\iota_j}\) over the stages, the total conditional failure probability over the stages that satisfy \eqref{eq:main-stage} is at most \(\delta/2\).

Put \(E_v=\Delta_v+\min\{d,w_H\}\excess_v\). If the improvement condition \(E_v\le c_*r_1^2\) fails at the first stage, then \(E_v\) is at least a constant and the error bound follows from \(\h^2\le2\). Otherwise, let \(j_*\) be the first stage whose radius violates \(E_v\le c_*r_j^2\), or the final stage if there is no such stage, and let \(a_{j_*}\) be the anchor at the start of stage \(j_*\). All earlier stages satisfy \eqref{eq:main-stage} and contract, so \(\h(Q_{a_{j_*}},Q_v)\le r_{j_*}\), and the definition of \(j_*\) gives
\[
                  r_{j_*}^2\le C E_v+c_\varepsilon^2\varepsilon .
\]
The algorithm does not need to identify \(j_*\). The final validation uses a new batch of size \(C\log(2/\delta)/\varepsilon\) and gives anchor \(a_j\) weight proportional to \(2^{-(J_\varepsilon-j)}\). The penalty for \(j_*\) is then at most \(C(\log(2/\delta)+J_\varepsilon-j_*+1)/n_{\rm val}\). Since the radii decrease geometrically, \(J_\varepsilon-j_*\le C\,r_{j_*}^2/(c_\varepsilon^2\varepsilon)\), so a large enough validation coefficient absorbs this penalty into \(CE_v\) plus a small multiple of \(\varepsilon\). The triangle inequality and \eqref{eq:validate} then bound the error of the output by \(\varepsilon+C E_v\) for this fixed \(v\), after rescaling the internal precision. A comparator that minimizes \(E_v\) exists by compactness and continuity; alternatively, one may fix an arbitrarily near-minimizing comparator before sampling. Since \(w_H=\sqrt H\), this is the error bound of Theorem~\ref{thm:proper}.

Finally, stage \(j\) costs \(O((dL_H+\min\{d,w_H\}\iota_j)/r_j^2)\) trajectories. Since \(r_j^{-2}\) grows by a factor four per stage, the stage \(k\) steps before the end carries weight of order \(4^{-k}\) in the total, which pays for the linear growth of \(\iota_j\) without a factor \(\log(1/\varepsilon)\). The total is \(O((dL_H+\min\{d,w_H\}\log(2/\delta))/\varepsilon)\). The design computations, the convex weight programs and the approximate projections run in polynomial time, the small-dimension enumeration is polynomial in \(H\), and every validation and prediction evaluates conditional models in time polynomial in the input parameters. The output is a single feasible parameter.

\section{Proof of Theorem~\ref{thm:lower}}
\label{app:lower}

We reduce from the gap version of Label Cover with perfect completeness. A bipartite graph has left vertices \(\mathcal U\) with alphabet \([L_1]\) and right vertices \(\mathcal V\) with alphabet \([L_2]\), and each edge \(e=(u,v)\) carries a projection \(\pi_e:[L_1]\to[L_2]\); a labeling satisfies \(e\) when the image of the left label equals the right label. For each fixed \(s_0>0\), it is NP-hard to distinguish graphs that admit a labeling satisfying every edge from graphs in which every labeling satisfies at most an \(s_0\) fraction of the edges, and for fixed \(s_0\) both alphabet sizes can be taken constant and the maps total \citep[Theorem~5]{bhangale26}.

\paragraph{Construction.}
Fix an integer \(m\ge2\) and use the soundness threshold \(s_0=1/(4m)\). The parameter set is the product of probability simplices, one per vertex, with coordinates \(\ell_{u,s}\) at a left vertex and \(r_{v,i}\) at a right vertex, so that \(d=L_1|\mathcal U|+L_2|\mathcal V|\). Each trajectory draws its context as in Section~\ref{sec:lower}: a uniform edge, independent uniform residues \(U_i\in\{0,\ldots,m-1\}\) indexed by the right labels, and a uniform sign \(b\in\{-1,1\}\). The model logit at every token is \(F_b/m\), and the target logit is \(b/m\). Under both the target and every model law, the \(H=C_0m^2\) tokens are conditionally independent given the context, where \(C_0\) is a large absolute integer.

The modular form of \(F_b\) serves two purposes. Because all residues lie in \(\{0,\ldots,m-1\}\), the model logit has magnitude less than one, which keeps the instance within the norm bounds. At the same time, shifting a residue by \(b\) is linear in the soft labels except on the wrap-around events \(U_{\pi_e(s)}=m-1\) for \(b=+1\) and \(U_{\pi_e(s)}=0\) for \(b=-1\), and these events have probability \(1/m\) for each label. A consistent one-hot labeling therefore reproduces the target logit except on a \(1/m\) fraction of contexts, whereas soft labels pay for the wrap-arounds in proportion to their mass, as the soundness argument shows.

The instance belongs to the model class of Section~\ref{sec:model}. At every step the contrast is the coefficient vector of \(F_b/m\), and the two action features are \(\pm\) one half of that vector. The entries of the features have magnitude at most \(1/2\), so each feature has Euclidean norm at most \(\sqrt d/2\), and the product of simplices has Euclidean radius at most \(\sqrt d\). Each per-action inner product has magnitude at most \(1/2\). Hence the instance satisfies \(B=\sqrt d\) and \(\Bdot=1\). Features are computable from the context and the graph, and projection onto \(\Theta\) projects each block onto its simplex by sorting and thresholding.

\paragraph{Completeness.}
Take the one-hot parameter given by a satisfying labeling. On every edge, the residue read by the left vertex through \(\pi_e\) is the residue of the right label, so the two logits coincide unless adding \(b\) wraps that residue around, an event of probability \(1/m\). The conditional squared Hellinger distance is at most two on every context, so \(\OPT\le2/m\).

\paragraph{Soundness.}
We show that on graphs of low value every feasible parameter has a logit error larger than \(a/m\) on a constant fraction of contexts, with \(a=1/16\) and \(\beta=1/200\). Fix a feasible parameter and suppose, for contradiction, that \(\Prob(|F_b-b|>a)<\beta\). By Markov's inequality, at least half of the edges have conditional failure probability at most \(1/100\). Fix one such edge, let \(\mu_i=\sum_{s:\pi_e(s)=i}\ell_{u,s}\) be the left soft labels pushed forward to the right labels, and let \(r_v=(r_{v,i})_i\) be the right soft labels. On the residue table of the context define
\[
 S=\sum_i(\mu_i-r_{v,i})U_i,\qquad
 W_+=m\sum_i \mu_i \ind_{\{U_i=m-1\}},\qquad
 W_-=m\sum_i \mu_i \ind_{\{U_i=0\}} .
\]
Here \(S\) compares the residues read by the two endpoints, and \(W_\pm\) is \(m\) times the left mass on labels whose residue wraps under the shift \(\pm1\). Expanding the modular arithmetic gives
\[
 F_+-1=S-W_+,\qquad F_-+1=S+W_- .
\]
Both signs act on the same residue table, and this is what makes the argument work. Since the sign is uniform, the failure probabilities of the two signs on this edge sum to at most \(2/100\), so with probability at least \(0.98\) over the table both signed errors have magnitude at most \(a\). Because \(W_\pm\ge0\), on this event \(|S|\le a\) and \(W=W_++W_-\le2a\). A single sign would only constrain the difference \(S-W_+\), whereas the two signs together constrain the wrap mass \(W\) and the moment \(S\) separately.

The wrap constraint forces the soft labels to concentrate. The two wrap events of one right label are disjoint and the residues are independent, so
\[
      \E W=2,\qquad \E W^2=4+(2m-4)\|\mu\|_2^2
                          \le 4+2m\|\mu\|_2^2 .
\]
Thus \(W\) has mean two for every choice of soft labels, but it is at most \(2a\le1\) with probability at least \(0.98\). Its mass above one must therefore come from a rare event, which requires a large second moment. Precisely, \(\E[W\ind_{\{W\le1\}}]\le1\), so the Cauchy--Schwarz inequality gives
\(1\le \E[W\ind_{\{W>1\}}]\le (0.02\,\E W^2)^{1/2}\),
and consequently \(m\|\mu\|_2^2\ge23\).

The moment constraint forces the two endpoints to agree. A centered uniform residue has second and fourth centered moments \((m^2-1)/12\) and \((m^2-1)(3m^2-7)/240\), so its fourth moment is at most three times its squared variance. Since \(\sum_i(\mu_i-r_{v,i})=0\), the sum \(S\) has mean zero and variance \((m^2-1)\|\mu-r_v\|_2^2/12\), and independence gives \(\E S^4\le3(\E S^2)^2\). If \(\E S^2>2a^2\), the Paley--Zygmund inequality applied to \(S^2\) would give \(\Prob(|S|>a)\ge1/12>0.02\), contradicting the event above. Therefore \(\|\mu-r_v\|_2\le 6a/m\) for \(m\ge2\), and since \(\|\mu\|_2\le1\),
\[
 \mu^\top r_v \ge\|\mu\|_2^2-\|\mu\|_2\|\mu-r_v\|_2
                \ge (23-6a)/m>1/m .
\]

Concentration and agreement together make randomized rounding effective. Choose the label of every vertex independently according to its soft labels. The selected edge is then satisfied with probability exactly \(\mu^\top r_v>1/m\), and since at least half of the edges are of this kind, the expected fraction of satisfied edges exceeds \(1/(2m)\). Some labeling therefore satisfies more than a \(1/(2m)\) fraction of the edges, which contradicts the soundness threshold \(1/(4m)\). Hence, on a graph of low value, every feasible parameter satisfies \(\Prob(|F_b-b|>a)\ge\beta\).

\paragraph{Detecting the error from one trajectory.}
The horizon \(H=C_0m^2\) is chosen so that a logit error of order \(1/m\) is visible in a single trajectory. Flag a trajectory if its fraction of ones differs from the known target mean \(\sigma(b/m)\) by more than \(a/(12m)\). Both logits lie in \([-1,1]\), where \(\sigma'>1/6\), so on the event \(|F_b-b|>a\) the model mean and the target mean differ by more than \(a/(6m)\). Conditional on the context, the fraction of ones is an average of \(H\) independent Bernoulli variables, so Hoeffding's inequality at \(H=C_0m^2\) shows that, for \(C_0\) large enough, a trajectory from the target is flagged with probability at most \(\beta/16\), and a trajectory from the model on such a context is flagged with probability at least \(1-\beta/16\). Consequently, on a graph of low value every feasible parameter is flagged with probability at least \(\beta(1-\beta/16)\). The same bound holds after averaging over any mixing law that is independent of the new context, and therefore for context-independent mixtures.

\paragraph{The reduction.}
Suppose that an algorithm as in Theorem~\ref{thm:lower} exists. For an output law \(R\) with \(\h^2(P,R)<(\beta/8)^2\), the flag probability is at most \(\beta/16+\beta/8<\beta/4\), because the total variation distance satisfies \(\TV(P,R)\le\h(P,R)\). On a satisfiable graph \(\sqrt H\,\OPT\le2\sqrt{C_0}\), so for \(\varepsilon_0\) and \(c\) small enough the guarantee of the algorithm yields such an output with probability at least \(9/10\). On a graph of low value, every output that is proper or a context-independent mixture is flagged with probability at least \(\beta(1-\beta/16)\). Using a fresh context and one rollout per test, a constant number of independent repetitions estimates the flag probability to within \(\beta/8\) with failure probability below \(1/10\), and comparing the estimate with \(\beta/2\) distinguishes the two cases with bounded error.

It remains to check that this procedure runs in randomized polynomial time. Fix \(m\), and hence \(H=C_0m^2\), before letting the graph grow, so that the alphabet sizes are constant; taking disjoint copies of the graph preserves its value and makes \(d\) arbitrarily large. The procedure simulates the trajectories from the target by drawing contexts and coins with the public mean \(\sigma(b/m)\), and it answers feature and projection queries from the graph, so the budget of the algorithm, including its oracle calls and the rollouts of its output, is polynomial in the size of the graph. The target coins and the test threshold can be computed with inverse-polynomial absolute error in the total simulation length; a telescoping coupling accounts for the finite precision of the target draws, and the exact probability thresholds have fixed slack at each fixed \(m\), so rounding preserves the gap. The procedure would therefore decide an NP-hard problem with bounded error, contradicting \(\mathrm{NP}\not\subseteq\mathrm{BPP}\). The separation constants, and hence \(\varepsilon_0\) and \(c\), do not depend on \(m\). Since \(m\) can be arbitrarily large, a guarantee whose factor is \(o(\sqrt H)\) uniformly over the dimension would fall below \(c\sqrt H\) at \(H=C_0m^2\) for some \(m\), which is excluded.

\end{document}